\documentclass[runningheads]{llncs}
\usepackage[T1]{fontenc}
\usepackage{multirow}
\usepackage{subcaption}
\usepackage{threeparttable}
\usepackage{caption}
\usepackage{hyperref}
\usepackage{amsmath}  
\usepackage{amsfonts} 
\usepackage{amssymb}  
\usepackage{algorithm}
\usepackage{booktabs} 
\usepackage{array}    
\usepackage{graphicx}
\usepackage{algpseudocode}
\algnewcommand\algorithmicinput{\textbf{Input:}}
\algnewcommand\algorithmicoutput{\textbf{Output:}}
\algnewcommand\Input{\item[\algorithmicinput]}
\algnewcommand\Output{\item[\algorithmicoutput]}
\begin{document}
\title{Noisy Deep Ensemble: Accelerating Deep Ensemble Learning via Noise Injection}
%
%
\author{Shunsuke Sakai\inst{1}\orcidID{0009-0008-4786-7679} \and
Shunsuke Tsuge\inst{1}\orcidID{0009-0003-8708-9094} \and
Tatsuhito Hasegawa\inst{1}\orcidID{0000-0002-0768-1406}}%
\authorrunning{F. Author et al.}

%
\institute{$^1$Graduate School of Engineering, University of Fukui\\
  3-9-1 Bunkyo, Fukui City, Fukui 910-8507, Japan, \\
  \email{mf240599@g.u-fukui.ac.jp}  
}
\maketitle              
\begin{abstract}
Neural network ensembles is a simple yet effective approach for enhancing generalization capabilities. The most common method involves independently training multiple neural networks initialized with different weights and then averaging their predictions during inference. However, this approach increases training time linearly with the number of ensemble members. To address this issue, we propose the novel ``\textbf{Noisy Deep Ensemble}'' method, significantly reducing the training time required for neural network ensembles. In this method, a \textit{parent model} is trained until convergence, and then the weights of the \textit{parent model} are perturbed in various ways to construct multiple \textit{child models}. This perturbation of the \textit{parent model} weights facilitates the exploration of different local minima while significantly reducing the training time for each ensemble member. We evaluated our method using diverse CNN architectures on CIFAR-10 and CIFAR-100 datasets, surpassing conventional efficient ensemble methods and achieving test accuracy comparable to standard ensembles. Code is available at \href{https://github.com/TSTB-dev/NoisyDeepEnsemble}{https://github.com/TSTB-dev/NoisyDeepEnsemble}

\end{abstract}

\section{Introduction}
\label{sec:Introduction}
Deep neural networks have achieved remarkable results in various fields, such as image recognition, natural language processing, and speech recognition. Their success can be attributed to their exceptionally high representation capacity. However, deep neural networks often lead to overfitting, particularly when the training data size is small. Consequently, the assessment of deep neural networks primarily focuses on their capability to effectively predict outcomes on new, unseen data. 

When multiple neural networks are trained independently, the randomness in the initialization of weights and the selection of mini-batches in SGD leads each network to learn different feature representations. Ensemble learning involves combining the predictions of multiple models to produce a more accurate prediction during inference. By integrating the predictions of multiple models, ensemble learning improves accuracy and uncertainty estimation performance compared to using a single model  \cite{DeepEnsembleLL,SimpleAS,WhyMH}. Ensemble learning is known to be more effective when each ensemble member possesses comparably high accuracy and makes mistakes on different samples  \cite{Snapshot}.

Ensemble learning is a simple yet effective approach for improving generalization performance, but it faces the issue of training time increasing linearly with the number of ensemble members. This issue is especially significant for deep neural networks, as training just one model can require weeks to months, which restricts the practical use of ensemble learning. Additionally, there are reported scaling laws indicating that the performance of neural networks improves with increased data size, model size, and computation  \cite{NeuralScalingLaw}. Furthermore, a phenomenon known as ``Grokking'', where continued training beyond the point of sufficient loss reduction can enhance generalization performance, has been observed in specific problems  \cite{AlwaysGrok,Grokking}. These factors contribute to the increased training time for a single model, making realizing their ensembles more challenging.

Several methods have been proposed to reduce the training time of ensemble learning\cite{Snapshot,GroupEnsemble,MotherNets,Dropout,DropConnect,PseudoEnsemble}. These methods improve the efficiency of ensemble learning by sharing parameters among ensemble members\cite{GroupEnsemble,DropConnect,Dropout}, constructing ensembles from the training process of a single model\cite{Snapshot}, and speeding up training through the pre-training of a base model\cite{LwF,MotherNets}. However, these methods have lower test accuracy than standard ensembles, resulting in significant performance differences.

In typical neural network training, randomness is generally limited to weight initialization and mini-batch selection in Stochastic Gradient Descent (SGD). However, additional perturbations can also be introduced. A notable example includes the addition of noise to inputs, which can enhance model regularization and robustness on out-of-distribution data  \cite{EffectOfNoise,RegularizingTE,NoiseIT}. There is also extensive research on perturbing the weights of neural networks  \cite{EffectOfNoise,NoisyTune,PseudoEnsemble,Flipout,WeightAug,SNIWD,ExplicitRI}. Weight perturbation can lead to effects such as regularization, exploration of parameter space, and robustness against adversarial perturbations. These approaches of introducing perturbations during neural network training are called Noise Injection. In this study, we specifically focus on perturbations to the weights.

We aim to reduce ensemble members' training time while achieving performance comparable to that of standard ensembles. In this study, we propose the Noisy Deep Ensemble, which utilizes noise injection to achieve this goal. In this method, a single model (\textit{parent model}) is trained until it converges, then perturbs its weights to construct multiple ensemble members (\textit{child models}). Since each ensemble member starts with reasonably good weights, they converge to nearby local minima after a short training period. Each ensemble member converges to different local minima compared to the original model, and their predictions exhibit diversity. Therefore, an ensemble of these models can achieve performance comparable to those trained independently at inference time.

\begin{figure*}[t]
\includegraphics[width=\textwidth]{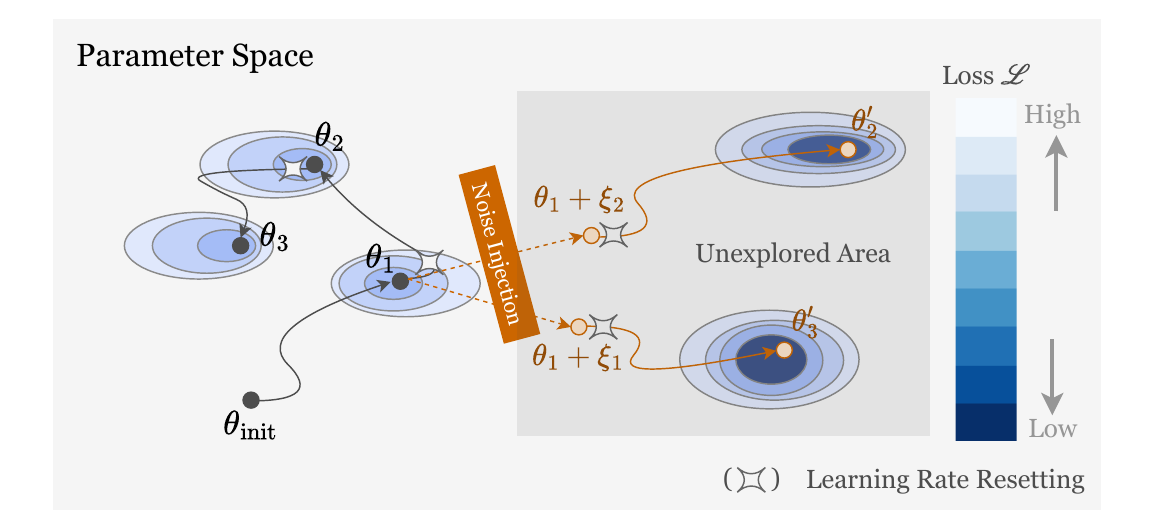}
\caption{Difference in the learning process between Noisy Deep Ensemble and the existing method (Snapshot Ensemble \cite{Snapshot}). Noisy Deep Ensemble promotes the exploration of wider parameter space by not being limited to the optimization path of SGD through Noise Injection. Snapshot Ensemble consists of 
$\mathcal{M} = \{\theta_2, \theta_3\}$, while Noisy Deep Ensemble consists of $\mathcal{M} = \{\theta_2', \theta_3'\}$.}
\label{fig:parameter_vis}
\end{figure*}

Fig.\ref{fig:parameter_vis} shows the differences in the learning process between the proposed method and the existing method (Snapshot Ensemble \cite{Snapshot}). The Snapshot Ensemble converges a single model to multiple local solutions by resetting to a high learning rate after convergence. On the other hand, the proposed method adds perturbations to the initial convergence point and explores a wider range of the parameter space, not limited to optimization by SGD.

The contributions of this study are outlined as follows:
\begin{enumerate}
    \item We developed a novel ensemble method called Noisy Deep Ensemble, which significantly reduces the training time of ensembles by perturbing the weights of neural networks.
    \item We evaluated the proposed method with various CNN architectures on CIFAR-10 and CIFAR-100 datasets. Noisy Deep Ensemble achieved higher test accuracies than the traditional efficient ensemble method, Snapshot Ensemble, across various CNN architectures.
    \item We conducted comprehensive experiments to investigate the diversity of predictions of each ensemble member and the impact of hyperparameters in the proposed method.
\end{enumerate}

\section{Background}
\label{sec:background}

\subsection{Ensemble Learning}
\label{subsec:ensemble_learning}

Here, we describe ensemble learning for neural networks in a standard class classification setting. Consider a given training set 
\(\mathcal{D}=\{(\boldsymbol{x}_i, y_i)\}_{i=1}^N\),
where \((\boldsymbol{x}_i, y_i)\) represents the pair of the \(i\)-th data point and its corresponding label. \(N\) denotes the size of the training data, and \(y\) belongs to \(\{1, 2, \dots, C\}\), with \(C\) being the number of classes. Let \(f_{\boldsymbol{\theta}}\) represent a neural network with parameters \(\boldsymbol{\theta}\). The optimization problem for a single neural network is formulated as follows:
\begin{equation}
\boldsymbol{\theta}^* = \underset{\boldsymbol{\theta}}{\operatorname{argmin}} \, \mathbb{E}_{(\boldsymbol{x}, y) \sim \mathcal{D}} \left[ \mathcal{L}(f_{\boldsymbol{\theta}}(\boldsymbol{x}), y) \right]
\label{eq:sgd}
\end{equation}
Here, \(\mathcal{L}\) is the loss function given the neural network's predicted probability distribution \(f_{\boldsymbol{\theta}}(\boldsymbol{x}) \in \mathbb{R}^C\) and the correct label \(y\), generally employing the cross-entropy loss function. In many cases, optimization, as depicted in Eq. (\ref{eq:sgd}), utilizes SGD.

In ensemble learning, \(M\) neural networks which have different initial weights, \(f_{\boldsymbol{\theta}^{(1)}}, f_{\boldsymbol{\theta}^{(2)}}, \dots, f_{\boldsymbol{\theta}^{(M)}}\), are independently optimized based on Eq. (\ref{eq:sgd}). At inference time, the ensemble prediction for a given test sample \(\boldsymbol{x}\) is determined by the average of the predictions from each neural network, denoted as \(P_{\text{ens}}\):
\begin{equation}
P_{\text{ens}} = \frac{1}{M} \sum_{j=1}^M f_{\boldsymbol{\theta}^{(j)}}(\boldsymbol{x})
\label{eq:ens_pred}
\end{equation}
When we mention ``standard ensemble'' in this paper, we refer to this particular form of ensemble learning.

\subsection{Efficient Ensemble Training}
\label{subsec:efficient_ensemble_training}

A key challenge in ensemble learning for neural networks is reducing training time. Standard ensemble learning increases training time in proportion to the number of ensemble members. Several methods have been proposed to reduce the training time of such ensemble learning  \cite{PseudoEnsemble,Snapshot,MotherNets}. The Snapshot Ensemble \cite{Snapshot} saves checkpoints from different learning stages of a single model and uses these models at each checkpoint for ensembling during inference. This allows ensemble learning to be accomplished within the training time of a single model. With regular learning rate scheduling, the predictions from these models at different checkpoints tend to be similar, limiting the effectiveness of ensemble learning. To address this issue, Snapshot Ensemble uses Cyclic Learning Rate Scheduling  \cite{CyclicLRSchedule} to encourage the model to converge different local minima, thus ensuring prediction diversity. However, its accuracy is lower compared to standard ensemble learning.

MotherNet \cite{MotherNets} involves pre-clustering multiple candidate model architectures and selecting a model called MotherNet within each cluster. MotherNet is trained using the entire dataset until convergence. After training, the MotherNet is expanded in depth and width using the technique proposed in Net2Net \cite{Net2Net}, creating multiple ensemble members known as ChildNets. The training time required for each ChildNet is significantly reduced compared to that of MotherNet. In MotherNet, noise from a Gaussian distribution is added to the weights of the trained MotherNet before constructing the ChildNets.

Group Ensemble \cite{GroupEnsemble} reduces the training time of each ensemble member by sharing the parameters of the lower layers among all ensemble members. Ensemble learning can be performed similarly to single-model training through Grouped Convolution \cite{AlexNet}. In addition, in ensemble learning with Knowledge Distillation \cite{KDHinton,LwF}, the base model is trained until convergence, and then the ensemble members, randomly initialized, are trained to mimic the output of the base model. Because the teacher signals from the base model provide richer label information, each ensemble member converges faster than regular training.

Methods such as Dropout \cite{Dropout} and DropConnect \cite{DropConnect} implicitly perform ensemble learning with multiple subnetworks. These techniques provide regularization effects and achieve higher test accuracy than a single model. Implicit ensemble learning requires approximately the same training time as a single model, but its test accuracy is lower than standard ensemble learning. The Pseudo Ensemble \cite{PseudoEnsemble} is a framework that generates multiple \textit{child model}s by perturbing a \textit{parent model}. DropConnect \cite{DropConnect} and Dropout \cite{Dropout} are special cases of \cite{PseudoEnsemble}.

Our proposed Noisy Deep Ensemble, similar to the Snapshot Ensemble \cite{Snapshot}, attempts to escape from local minima by resetting the learning rate. Unlike the Snapshot Ensemble, the Noisy Deep Ensemble additionally encourages the exploration of the parameter space through perturbations to the weights, enhancing the diversity of predictions in ensemble learning. Additionally, unlike MotherNet \cite{MotherNets}, the Noisy Deep Ensemble maintains the same model architecture across all ensemble members, making it simpler and easier to implement. And also, MotherNet makes little mention of weight perturbation. The Noisy Deep Ensemble can be considered a form of pseudo-ensemble \cite{PseudoEnsemble} learning. However, it is not an implicit form of ensemble learning that uses a single model but an explicit form that utilizes multiple models for inference, achieving higher test accuracy.

\subsection{Noise Injection}
\label{subsec:noise_injection}

Noise Injection is a technique for introducing some form of perturbation during neural network training to enhance generalization performance and robustness against perturbations. Various forms of perturbations can be considered, including perturbations to the inputs, the model’s outputs, or the weights. In this paper, we focus on perturbations to the weights.

In NoisyTune \cite{NoisyTune}, noise from a uniform distribution is added once to the weights of a pre-trained language model before fine-tuning for downstream tasks. Adaptively varying the strength of the noise according to the standard deviation of the weights improves performance in downstream tasks. This approach perturbs the weights to mitigate overfitting of the language model to the pre-training tasks.
In WeightAugmentation \cite{WeightAug}, random rigid transformations are applied to the weight matrices of neural networks during training. This acts as a form of regularization and has been shown to improve test accuracy across various CNN architectures. 
DropConnect \cite{DropConnect} and other forms of pseudo-ensemble learning \cite{PseudoEnsemble} also involve perturbing weights, contributing to its regularization effects.

An \cite{EffectOfNoise} demonstrates, under certain assumptions, the theoretical impact of perturbations to model weights on learning. When weight noise is sampled from a Gaussian or uniform distribution and is of a sufficiently small scale, this additive noise can make neural networks more sparse and suppress overfitting. On the other hand, these results are based on an assumption continuously perturbing the weights during training.

Bayesian neural networks learn the posterior distribution of weights over a given training set rather than providing point estimates of optimal weights. Variational Bayesian neural networks approximate this posterior distribution of weights using variational inference to minimize the Evidence Lower BOund (ELBO) \cite{VariationalBNN}. Here, perturbations of weights can also be considered as sampling from some posterior distribution of weights. While Bayesian neural networks tend to explore a single mode in the function space, ensemble learning explores multiple modes \cite{DeepEnsembleLL}. Therefore, generally, ensemble learning exhibits superior generalization performance and uncertainty estimation compared to Bayesian neural networks.

The model weights are perturbed only once after \textit{parent model} training in the Noisy Deep Ensemble to explore parameter space effectively. Then, multiple \textit{child models} are instantiated with different perturbed weights. This differentiates it from approaches \cite{EffectOfNoise,PseudoEnsemble,WeightAug} that continuously perturb the weights during training. The purpose of perturbation in the Noisy Deep Ensemble is similar to that of NoisyTune \cite{NoisyTune}, aiming to explore the parameter space and encourage convergence to different local minima. Additionally, the Noisy Deep Ensemble incorporates the advantages of both Bayesian neural networks and ensembles, exploring diverse modes within the function space while accounting for the uncertainty of the weights.

\section{Noisy Deep Ensemble}
\label{sec:noisy_deep_ensemble}

\begin{figure*}[t]
\includegraphics[width=\textwidth]{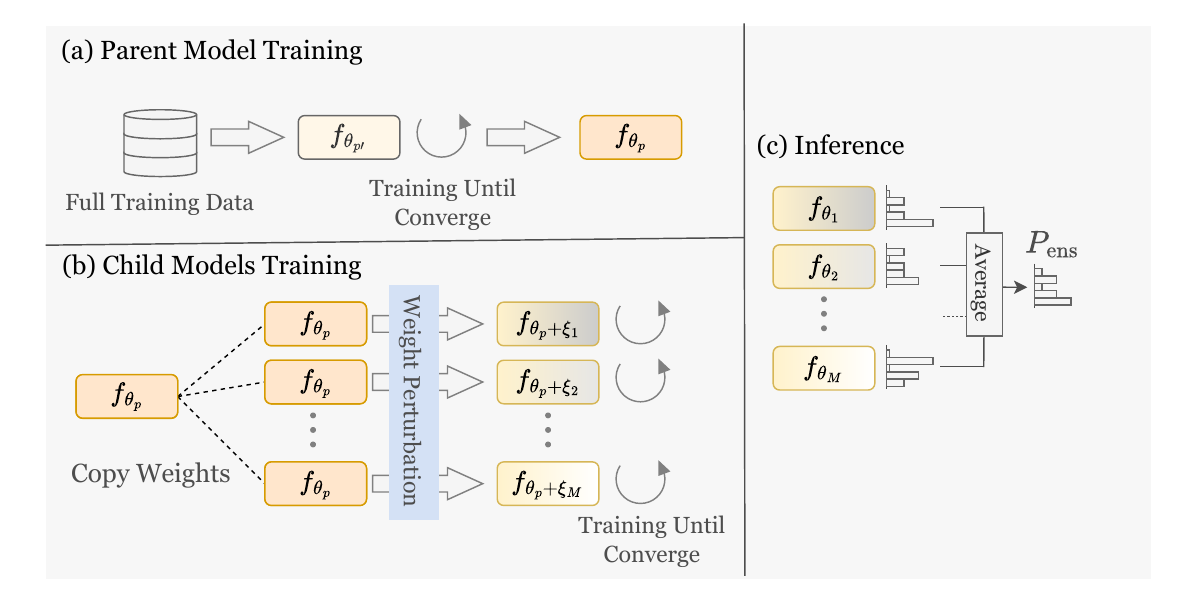}
\caption{Overview of the Noisy Deep Ensemble}
\label{fig:method}
\end{figure*}

Fig.\ref{fig:method} provides an overview of the proposed method: Noisy Deep Ensemble. The training process of the Noisy Deep Ensemble consists of two stages: training the \textit{parent model} and training the\textit{child models}. First, the \textit{parent model} is trained until it converges on all available training data(Fig.\ref{fig:method}(a)). Afterward, the weights of the trained \textit{parent model} are duplicated to construct each ensemble member (\textit{child models}). The \textit{child models} have the same network architecture as the \textit{parent model}, and their initial weights are identical to those of the trained \textit{parent model}. Therefore, simply retraining the \textit{child models} with standard SGD does not produce sufficient diversity in the ensemble predictions. To address this, we perturb the weights of each \textit{child model} to encourage exploration of different local minima. After perturbing the weights of the \textit{child models}, they are independently trained until convergence (Fig.\ref{fig:method}(b)). During inference, the ensemble's prediction is obtained by averaging the predicted probability distributions of all \textit{child models} (Fig.\ref{fig:method}(c)).

We present the Noisy Deep Ensemble details, considering a standard classification setting. Let us assume the training data set is given as follows:
\begin{equation}
\mathcal{D_\text{train}} = \{(\boldsymbol{x}_i, y_i)\}_{i=1}^N,
\end{equation}
where $(\boldsymbol{x}_i, y_i)$ represents the data point and corresponding class label of the $i$-th sample, and $i \in \{1, 2, \ldots, N\}$, with $C$ being the number of classes. We define a neural network $f_{\boldsymbol{\theta}^{(p)'}}$ with initial weight $\boldsymbol{\theta}^{(p)'}$ as the \textit{parent model}, and train it using SGD according to Eq. (\ref{eq:sgd}) until convergence, where $\mathcal{D} = \mathcal{D_\text{train}}$. The weight of the trained \textit{parent model} is denoted as $\boldsymbol{\theta}^{(p)} \in \mathbb{R}^D$ where $D$ is the number of dimensions of weights.

We independently sample $M$ times from a noise distribution $p(\boldsymbol{\xi})$ to obtain $M$ noise vectors $\{\boldsymbol{\xi}^{(1)}, \boldsymbol{\xi}^{(2)}, \ldots, \boldsymbol{\xi}^{(M)}\}$. For any $i$, $\boldsymbol{\xi}^{(i)} \in \mathbb{R}^D$ can be added to the weights of the trained \textit{parent model}. Thus, we initialize the weights of multiple ensemble members (\textit{child models}) by applying different perturbations to the pre-trained \textit{parent model} as shown in Eq. (\ref{eq:noise_all}):
\begin{equation}
    \boldsymbol{\theta}^{(i)'} = \boldsymbol{\theta}^{(p)} + \boldsymbol{\xi}^{(i)}, \quad i = 1, 2, \dots, M
    \label{eq:noise_all}
\end{equation}
Although Eq. (\ref{eq:noise_all}) perturbs all weights, in this study, we selectively perturb weights by using a random mask vector $\boldsymbol{m}^{(i)} \in \{0, 1\}^D$ as follows:
\begin{equation}
    \boldsymbol{\theta}^{(i)'} = \boldsymbol{\theta}^{(p)} + \boldsymbol{\xi}^{(i)} \odot \boldsymbol{m}^{(i)}, \quad i = 1, 2, \dots, M
    \label{eq:noise_masked}
\end{equation}
where $\odot$ denotes the hadamard product. Eq. (\ref{eq:noise_all}) is a special case in Eq. \ref{eq:noise_masked} where the weight perturbation by $\boldsymbol{m}^{(i)} \in [1, 1, \dots, 1]^\top$.

Subsequently, each \textit{child model} is trained using SGD on $\mathcal{D_\text{train}}$ until convergence, following Eq. (\ref{eq:sgd}). As each \textit{child model} is initialized using the weights of the pre-trained \textit{parent model}, they converge more quickly. Increasing the perturbation scale leads to a more significant deviation from the pre-trained weights of the \textit{parent model}, resulting in a longer training time for the \textit{child models}. Conversely, decreasing the perturbation scale tends to result in convergence to the same local minima as the pre-trained \textit{parent model}, reducing the diversity of predictions in ensemble learning. Thus, the perturbation scale balances a trade-off between the diversity of predictions in ensemble learning and training time.

During inference, the average prediction probability distribution of the trained \textit{child models} \( f_{\boldsymbol{\theta}^{(1)}}, f_{\boldsymbol{\theta}^{(2)}}, \dots, f_{\boldsymbol{\theta}^{(M)}} \) is computed as Eq. (\ref{eq:ens_pred}), and this average is used as the ensemble's overall prediction. Each \textit{child model} converges to different local minima through their weights perturbations and short-time training, providing diversity in their predictions. As a result, the ensemble can achieve superior test accuracy compared to individual \textit{child models} or the \textit{parent model} alone.

The key hyperparameters in a Noisy Deep Ensemble are the proportion of weights subject to perturbation and the scale of perturbation. The proportion of weights to be perturbed, denoted as \(\alpha\), is related to the mask vector \(\boldsymbol{m}\), and is given by \(\alpha = \frac{1}{D}\sum_{i=1}^D m_i\). More precisely, each element of \(\boldsymbol{m}\) is sampled from Bernoulli distribution, i.e., $m_i \sim \text{Bernoulli}(p)$, and $p$ is equivalent to $\alpha$.
In this study, we consider using a Gaussian distribution \(\xi_i \sim \mathcal{N}(0, \beta) \) or a uniform distribution \(\xi_i \sim U(-\beta, \beta)\) as the perturbation distribution, where \(\beta\) represents the scale of the perturbation.

\section{Experiments}
\label{sec:experiments}

\subsection{Experiment Setting}
\label{subsec:experiment_setting}

In this study, we validate the effectiveness of our proposed method using CIFAR-10 (C10) and CIFAR-100 (C100) \cite{CIFAR}. 
As evaluation metrics, we employ test accuracy on CIFAR-10/100. To demonstrate the effectiveness of our method across various CNN architectures, we utilize popular CNN architectures such as ResNet18 \cite{ResNet}, VGG16 \cite{VGG}, and EfficientNetB0 \cite{EfficientNet}.

During training, the mini-batch size is set to 64, and Momentum SGD \cite{Momentum} is used as the optimizer, with a momentum value of 0.9 and weight decay set to 0.0005. Learning rate scheduling employs Cosine Scheduling \cite{CyclicLRSchedule}, with maximum and minimum learning rates set to 0.1 and 0.0, respectively. Throughout all experiments, the \textit{parent model} is trained for 200 epochs, while \textit{child models} are trained for 50 epochs. Similarly, single models and conventional ensemble learning are trained for 200 epochs unless specified. The number of ensemble members is set to 10 unless specified. When perturbing the weights, two hyperparameters are considered: the proportion of weights perturbed, $\alpha$, and the scale of perturbation, $\beta$. This study determined their optimal values through grid search, presented in Table \ref{tab:optimal_noise_values}. Unless specified, these values are used. The impact of these hyperparameters on test accuracy is evaluated in Section \ref{subsec:ablation_weight_pertubation}.

\begin{table}[]
\centering
\setlength{\tabcolsep}{4pt} 
\captionsetup{skip=2pt}
\caption{Optimal weight perturbation values found by grid search. $\alpha$ denotes the ratio of weights which are pertubated. $\beta$ denotes the strength of the noise.}
\label{tab:optimal_noise_values}
\begin{tabular}{@{}ccccccccccccc@{}}
\toprule
\multirow{3}{*}{Noise Parameter} & \multicolumn{4}{c}{ResNet18}                       & \multicolumn{4}{c}{VGG16}                          & \multicolumn{4}{c}{EfficientNetB0}                 \\ \cmidrule(l){2-13} 
                                 & \multicolumn{2}{c}{C10} & \multicolumn{2}{c}{C100} & \multicolumn{2}{c}{C10} & \multicolumn{2}{c}{C100} & \multicolumn{2}{c}{C10} & \multicolumn{2}{c}{C100} 
        \\ \cmidrule(l){2-13} 
                                 & $\alpha$          & $\beta$          & $\alpha$            & $\beta$           & $\alpha$          & $\beta$           & $\alpha$            & $\beta$           & $\alpha$            & $\beta$          & $\alpha$            & $\beta$  \\        \midrule 
$\xi_i \sim U(-\beta,\beta)$                              & 0.8        & 1.6        & 0.8         & 1.6        & 0.1        & 0.1        & 0.05        & 0.1        & 0.7         & 0.1       & 0.3         & 0.3        \\
$\xi_i \sim \mathcal{N}(0, \beta)$                             & 0.8        & 1.6        & 0.8         & 1.6        & 0.1        & 0.01       & 0.05        & 0.05       & 0.7           & 0.1        & 0.3         & 0.3     \\
\bottomrule
\end{tabular}
\end{table}

\subsection{Main Results}
\label{subsec:results}
To verify the effectiveness of the proposed method, we compared the test accuracies on CIFAR-10 and CIFAR-100 under the following settings:
\begin{itemize}
    \item \textbf{Single}: Single model.
    \item \textbf{Ensemble}: Standard ensemble learning, training $M$ models independently from different initial weights.
    \item \textbf{Noisy Single}: Single model in which weights were perturbed after first convergence and trained again until convergence. The perturbation distribution $p(\boldsymbol{\xi})$ includes both uniform (uni) and Gaussian (norm) distributions.
    \item \textbf{Noisy Ensemble}: Trained single model until convergence, then weights were perturbed, and multiple models were independently trained. Each model's weight noises are sampled independently from uniform (uni) or Gaussian (norm) distributions.
\end{itemize}

\begin{table}[t]
\centering
\setlength{\tabcolsep}{6pt} 
\captionsetup{skip=2pt}
\caption{Comparison of model performances on CIFAR10(C10) and CIFAR100 (C100) datasets. The highest accuracy is shown in \textbf{bold}, while the second highest is \underline{underlined}.}
\label{tab:model_performance}
\begin{tabular}{@{}lcccccc@{}}
\toprule
Method        & \multicolumn{2}{c}{ResNet18} & \multicolumn{2}{c}{VGG16} & \multicolumn{2}{c}{EfficientNetB0} \\ \cmidrule(l){2-7} 
              & C10      & C100     & C10      & C100   & C10        & C100       \\ \midrule
Single        & 0.8965       & 0.7350        & 0.8823       & 0.6023     & \underline{0.9186}         & 0.7197         \\
Ensemble      & \textbf{0.9158}       & \textbf{0.7739}       & \underline{0.9036}       & 0.6532     & \textbf{0.939 }         & \textbf{0.7782}         \\
Noisy Single (uni)  & 0.8863       & 0.7075       & 0.8936       & 0.6163     & 0.903          & 0.6929         \\
Noisy Single (norm) & 0.8871       & 0.7640        & 0.8893       & 0.6112     & 0.8999              & 0.6852         \\
Noisy Ensemble (uni)& \underline{0.9147}       & 0.7151       & \textbf{0.9072}       & \textbf{0.662}      & 0.9115         & \underline{0.7284}         \\
Noisy Ensemble (norm)& 0.9132      & \underline{0.7660}        & 0.9008       & \underline{0.6577}     & 0.9129              & 0.7178         \\ \bottomrule
\end{tabular}
\end{table}

Table \ref{tab:comparison_ensemble} presents the test accuracy in different ensemble configurations for CIFAR-10 and CIFAR-100. Standard ensemble learning (Ensemble) significantly improves test accuracy in all cases compared to a single model (Single). On the other hand, the Noisy Single approach, which involves training a single model with perturbed weights, does not consistently show superiority over the single model, and in some cases, test accuracy decreases depending on the model and dataset. This is true even when the type of perturbation distribution is varied. However, our Noisy Ensemble approach, which involves retraining multiple models with perturbed weights, surpasses the single model test accuracy in almost all configurations and demonstrates performance comparable to that of standard ensemble learning.

Table \ref{tab:comparison_ensemble} compares the performance with existing ensemble learning methods. The CIFAR-10 and CIFAR-100 datasets and the ResNet18 model architecture are used. The number of ensemble members is set to 10. The proposed method significantly outperforms all existing ensemble learning methods \cite{Snapshot,KDHinton,GroupEnsemble} and narrows the performance gap with standard ensembles. Knowledge Distillation Ensemble  \cite{KDHinton}, similar to Noisy Deep Ensemble, pre-trains a \textit{parent model}, but the weights of the \textit{child models} are randomly initialized. Therefore, it is possible that the models may not fully converge in a short training time.

\begin{table}[t]
\centering
\setlength{\tabcolsep}{6pt} 
\captionsetup{skip=2pt}
\caption{Comparison with other ensemble methods. KDE stands for Knowledge Distillation Ensemble \cite{KDHinton}, GE stands for Group Ensemble \cite{GroupEnsemble}, SE stands for Snapshot Ensemble \cite{Snapshot}, and BE stands for BatchEnsemble \cite{BatchEnsemble}.}
\label{tab:comparison_ensemble}
\begin{tabular}{@{}cccccc|c@{}}
\toprule
Method & KDE \cite{KDHinton} & GE \cite{GroupEnsemble} & SE \cite{Snapshot} & BE \cite{BatchEnsemble} & Ours & Standard \\ 
\midrule 
C10    & 0.8725              & 0.9043                 & 0.9088             & 0.8259                      & \underline{0.9132} & \textbf{0.9158} \\
C100   & 0.6771              & 0.6524                 & 0.7041             & 0.5869                      & \underline{0.7660} & \textbf{0.7739} \\
\bottomrule
\end{tabular}
\end{table}

\begin{table}[b]
\centering
\setlength{\tabcolsep}{8pt}
\captionsetup{skip=2pt}
\caption{Test accuracy of each ensemble member. }
\label{tab:single_accuracy}
\begin{tabular}{@{}lcccccc@{}}
\toprule
\textbf{Accuracy} & Child1 & Child2 & Child3 & Child4 & Child5 & Ensemble \\ \midrule
$\xi_i \sim U(-\beta,\beta)$ & 0.8836 & 0.8849 & 0.8911 & 0.8843 & 0.8876 & \textbf{0.9116} \\
$\xi_i \sim \mathcal{N}(0, \beta)$ & 0.8897 & 0.8884 & 0.8894 & 0.8861 & 0.8829 & 0.9102 \\ \bottomrule
\end{tabular}
\end{table}

\subsection{Evaluating Ensemble Effectiveness}
\label{subsec:evaluating_ensemble_effectiveness}
Ensemble learning is more effective when these two conditions are met: (i) each ensemble member exhibits high test accuracy, and (ii) ensemble members don't share miss-classified samples with each other \cite{Snapshot}. This section discusses about these conditions, (i) and (ii), in detail. We use the CIFAR-10 dataset and the ResNet18 model architecture in the subsequent experiments. The number of ensemble members is set to 5. The weights perturbation scale is the same as described in Section \ref{subsec:experiment_setting}.

Table \ref{tab:single_accuracy} shows the test accuracy of individual ensemble members in the Noisy Deep Ensemble. The test accuracy is sufficiently high for both uniform and Gaussian perturbation distributions, although they are slightly lower than those of the single models. For reference, the accuracy of a standalone ResNet18 on CIFAR-10 is 0.8965, as shown in Table \ref{tab:model_performance}. Furthermore, the performance is improved by their ensemble.

\begin{figure}[t]
    \centering
    \begin{subfigure}[b]{0.45\textwidth}
        \centering
        \includegraphics[width=\textwidth]{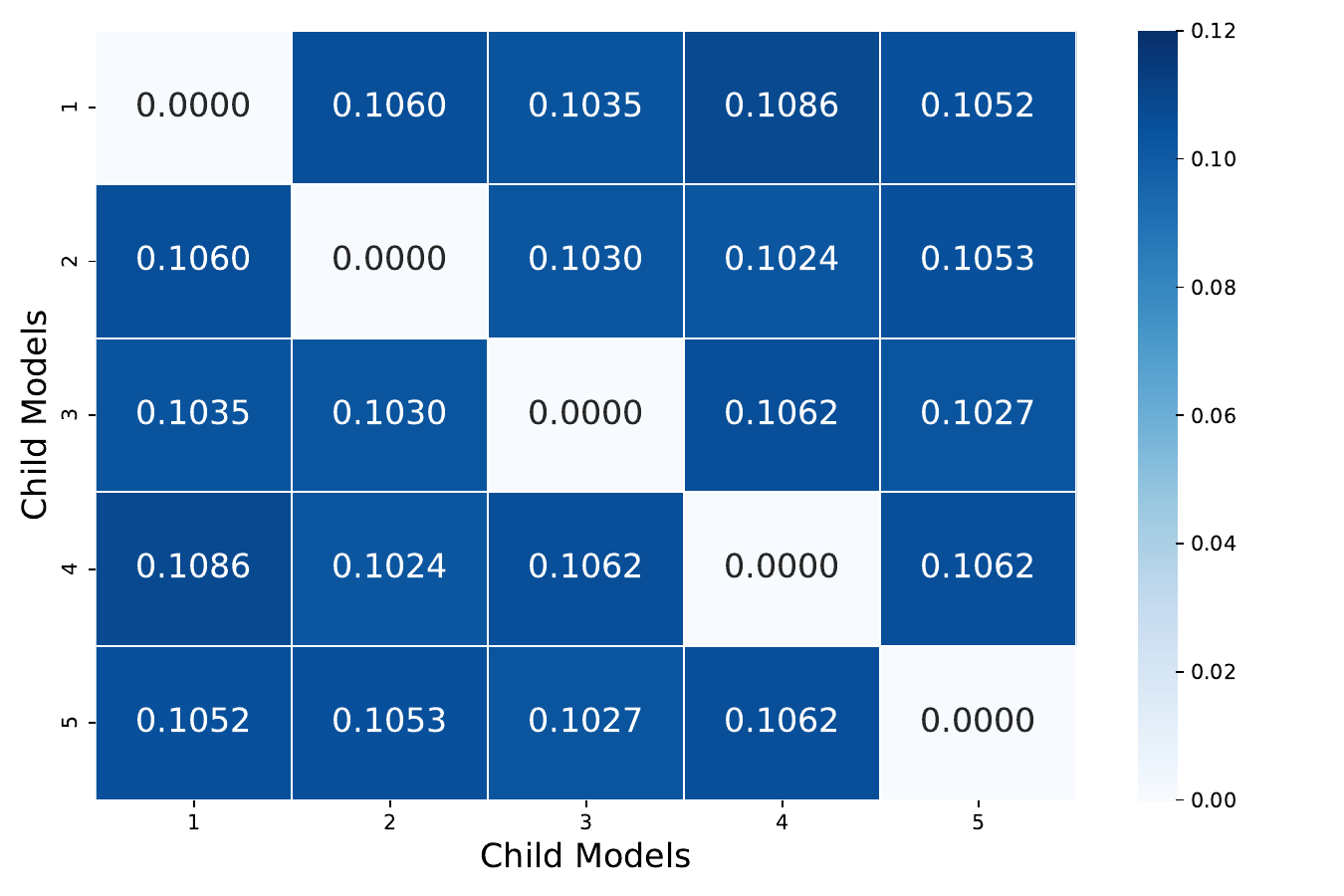}
        \caption{Noisy Deep Ensemble (uni)}
        \label{fig:disag_uni}
    \end{subfigure}
    \hfill
    \begin{subfigure}[b]{0.45\textwidth}
        \centering
        \includegraphics[width=\textwidth]{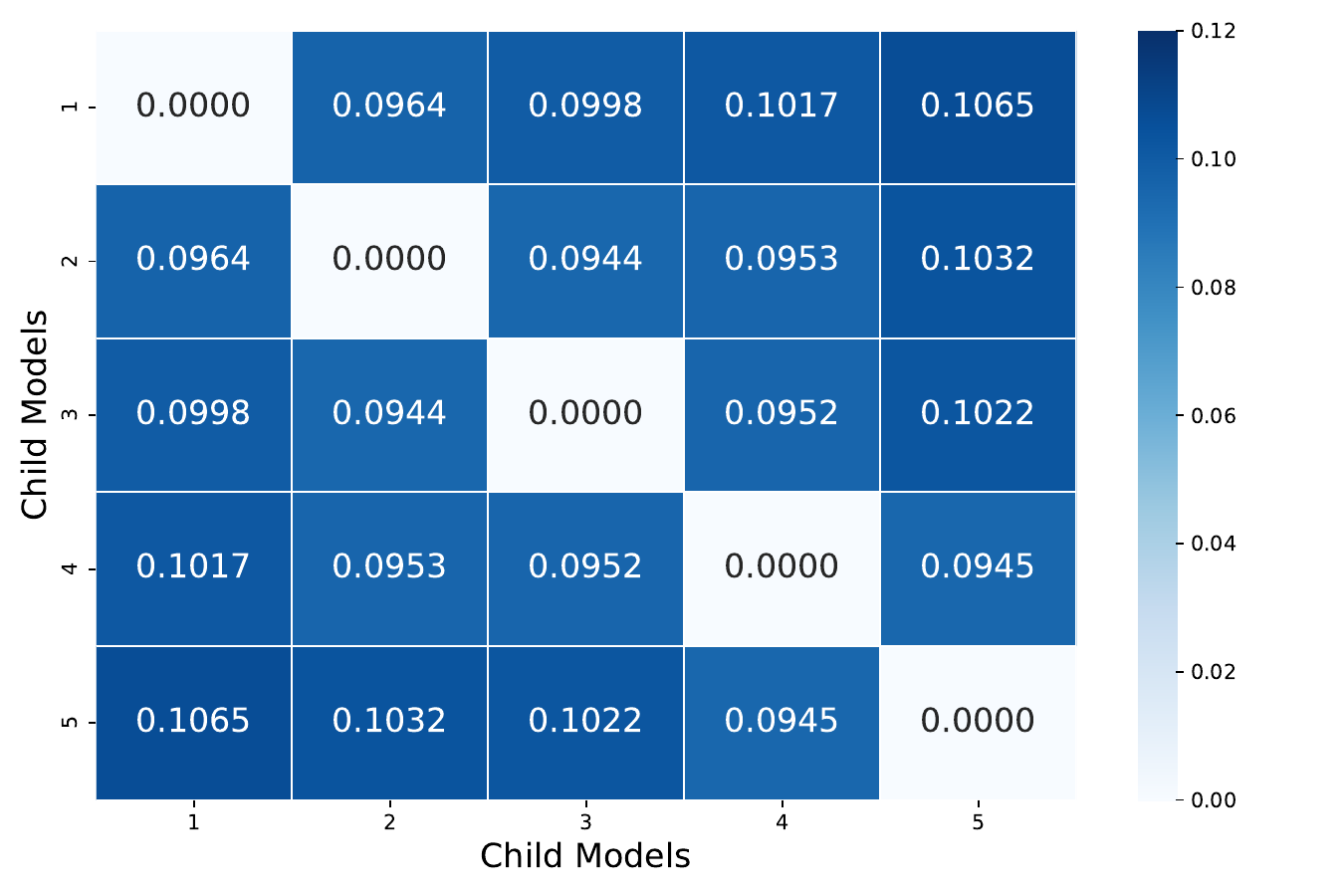}
        \caption{Snapshot Ensemble}
        \label{fig:disag_snap}
    \end{subfigure}
    \caption{The disagreement rate among ensemble members' predictions}
    \label{fig:disag}
\end{figure}

\begin{figure}[t]
    \centering
    \begin{subfigure}[b]{0.45\textwidth}
        \centering
        \includegraphics[width=\textwidth]{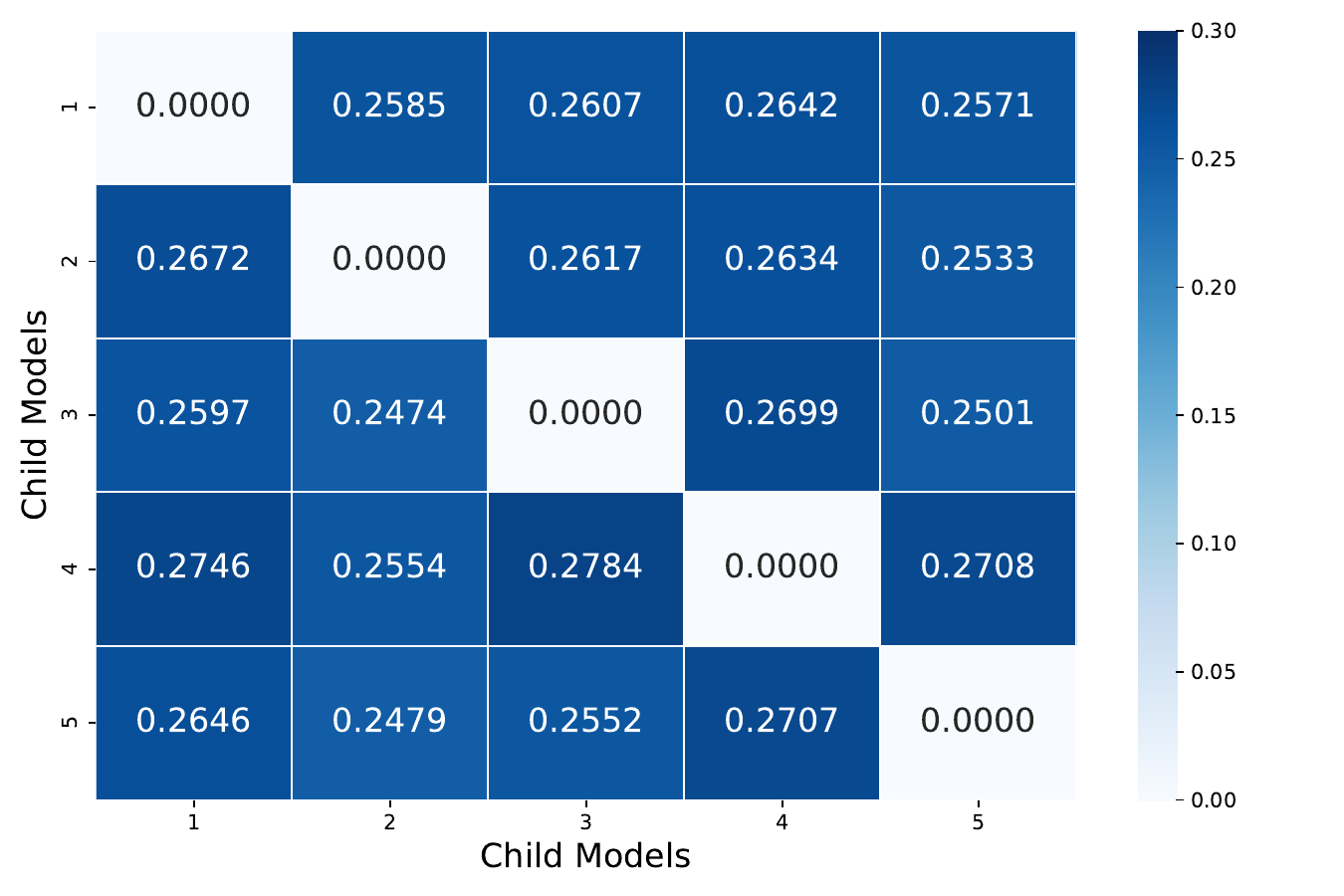}
        \caption{Noisy Deep Ensemble (uni)}
        \label{fig:kl_div_uni}
    \end{subfigure}
    \hfill
    \begin{subfigure}[b]{0.45\textwidth}
        \centering
        \includegraphics[width=\textwidth]{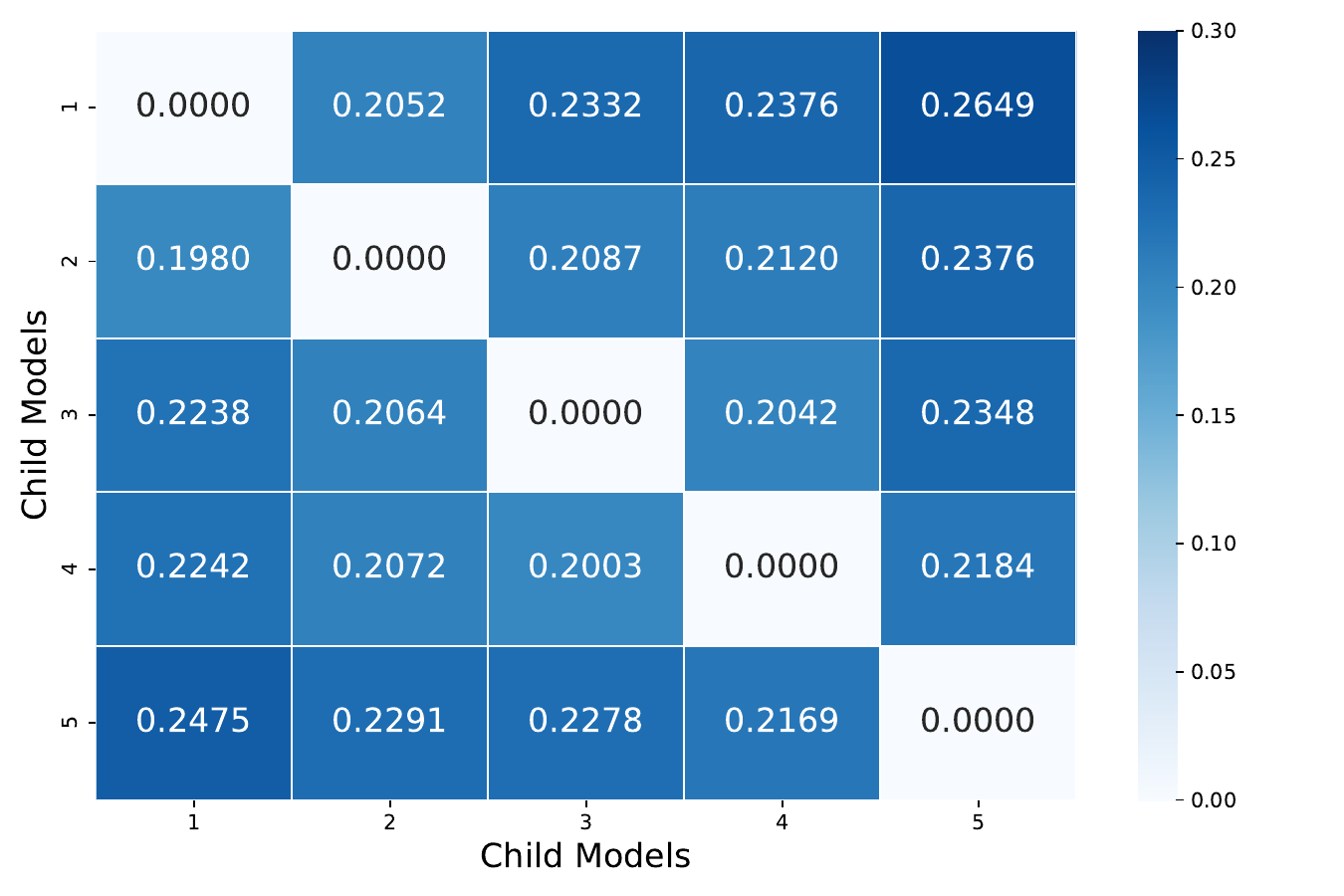}
        \caption{Snapshot Ensemble}
        \label{fig:kl_div_snap}
    \end{subfigure}
    \caption{Average KL divergence between the prediction probability distributions of ensemble members on the test data}
    \label{fig:kl}
\end{figure}


In ensemble learning, more significant benefits are obtained when the predictions of each ensemble member are diverse. One straightforward metric for evaluating such prediction diversity is the disagreement rate. Fig. \ref{fig:disag} shows the disagreement rates in predictions on the CIFAR-10 test data for both the Noisy Deep Ensemble and the Snapshot Ensemble. A uniform distribution (uni) is used as the perturbation distribution, and the number of ensemble members is set to 5. The disagreement rate is calculated for every pair of ensemble members. Compared to the Snapshot Ensemble, the Noisy Deep Ensemble exhibits a slightly higher rate of prediction disagreement, i.e., the Noisy Deep Ensemble has higher prediction diversity than the Snapshot Ensemble.

The KL divergence between the prediction probability distributions of each ensemble member also represents the diversity of predictions well. Fig. \ref{fig:kl} shows the average KL divergence of the prediction probability distributions on the CIFAR-10 test data for both the Noisy Deep Ensemble and the Snapshot Ensemble. Clearly, the Noisy Deep Ensemble exhibits a higher average KL divergence among ensemble members than the Snapshot Ensemble. The Noisy Deep Ensemble enhances the diversity of predictions and improves the test accuracy of the ensemble through perturbations to the weights and independent training of each ensemble member.
 



\subsection{Training Efficiency of Noisy Deep Ensemble}
\label{subsec:training_efficiency_analysis}

In this section, we examine the training time of the Noisy Deep Ensemble in comparison to standard ensemble learning. In standard ensemble learning, if the training time for a single model is $T_\text{single}$ and the number of ensemble members is $M$, the total training time is $T_\text{standard} = M \cdot T_\text{single}$. On the other hand, the total training time for the Noisy Deep Ensemble is $T_\text{parent} + T_\text{child} \cdot M$. Assuming that the training time for the \textit{child model} can be significantly reduced compared to the \textit{parent model}, with $T_\text{parent} \gg T_\text{child}$, the Noisy Deep Ensemble shows a linear reduction in training time relative to the number of ensemble members compared to standard ensembles. Therefore, the Noisy Deep Ensemble is more suitable for scaling the number of ensemble members than traditional ensemble learning. The impact of scaling the number of ensemble members on accuracy is examined in Section. \ref{subsec:scaling_ensemble_members}.
We compared the training times of standard ensemble learning and Noisy Deep Ensemble with different numbers of ensemble members using wall-clock time\footnote{The experiments were conducted using an NVIDIA GeForce RTX3090 (24GB) and an Intel Core i9-10850K @ 3.60GHz (32GB).}. The results are shown in Table \ref{tab:training_time}. Compared to the training time of a standard ensemble, the training time reduction of Noisy Deep Ensemble ($T_\text{noisy}$) decreases as the number of ensemble members increases. Notably, when the number of ensemble members is 10, the training time for the Noisy Deep Ensemble is 35\% of that for the standard ensemble. Additionally, compared to a standard ensemble, the degradation in test accuracy is minimized\footnote{We used uniform noise for weight perturbation as section \ref{subsec:evaluating_ensemble_effectiveness}.}.
\begin{table}[t]
\centering
\setlength{\tabcolsep}{8pt}
\captionsetup{skip=2pt}
\caption{Comparison of training time between standard and noisy deep ensemble models in wall-clock hours}
\begin{tabular}{@{}cccccc@{}}
\toprule
\# Ensemble members           & 2    & 4    & 6    & 8    & 10   \\ \hline
$T_\text{standard}$  [h]  & 2.1  & 4.2  & 6.3  & 8.3  & 10.4 \\ 
$T_\text{noisy}$  [h]& 1.6  & 2.1  & 2.6  & 3.1  & 3.6  \\ 
$T_\text{noisy}/{T_\text{standard }}$ [\%]  &  73\% & 50\% & 41\% & 37\% & 35\% \\
Accuracy Drop [$\Delta$] & 0.0058 & 0.0029 & 0.0013 & 0.0019 & 0.0025
\\
\bottomrule
\end{tabular}
\label{tab:training_time}
\end{table}


\subsection{Performance of Robustness and Calibration}
We evaluate the robustness of data corruption on CIFAR-10-C \cite{CIFAR-C}. The results are shown in Table \ref{tab:performance_cifar10_severity}. Noisy Deep Ensemble outperforms the single model across all corruption severity and performs similarly to the standard ensemble. This indicates the model's robustness can be improved by the noisy, deep ensemble. 

\begin{table}[t]
\centering
\setlength{\tabcolsep}{6pt} 
\captionsetup{skip=2pt}
\caption{Comparison of model performance on CIFAR-10 across different severity levels of degradation. The values in the table represent test accuracy.}
\label{tab:performance_cifar10_severity}
\begin{tabular}{@{}lcccccc@{}}
\toprule
Method        & \multicolumn{6}{c}{Severity} \\ \cmidrule(l){2-7} 
              &  0  & 1     & 2     & 3     & 4     & 5     \\ \midrule
Standard      & \textbf{0.9158}   & \textbf{0.8406} & \textbf{0.7846} & \textbf{0.7313} & \textbf{0.6648} & \textbf{0.5605} \\
Single        & 0.8965   & 0.8135 & 0.7549 & 0.7013 & 0.6379 & 0.5351 \\
NDE (uni)     & \underline{0.9147}  & 0.8262 & 0.7712 & 0.7172 & 0.6483 & 0.5456 \\
NDE (norm)    & 0.9132 & \underline{0.8281} & \underline{0.7743} & \underline{0.7218} & \underline{0.6531} & \underline{0.5509} \\ \bottomrule
\end{tabular}
\end{table}

Table \ref{tab:calib_performance_cifar10_cifar100} shows the calibration performance on CIFAR-10 and CIFAR-100. Noisy Deep Ensemble is better overall than the single model regarding calibration performance. However, the difference in calibration performance between the standard ensemble and the noisy deep ensemble is significant compared to the difference in accuracy. The improvement of Noisy Deep Ensemble performance in calibration remains as future work.

\begin{table}[t]
\centering
\setlength{\tabcolsep}{6pt} 
\captionsetup{skip=2pt}
\caption{Comparison of model performance on CIFAR-10 and CIFAR-100 datasets. Acc. represents test accuracy (↑), ECE is expected calibration error (↓), and NLL is negative log-likelihood (↓).}
\label{tab:calib_performance_cifar10_cifar100}
\begin{tabular}{@{}lcccccc@{}}
\toprule
\multirow{2}{*}{Method} & \multicolumn{3}{c}{CIFAR-10}    & \multicolumn{3}{c}{CIFAR-100} \\ \cmidrule(lr){2-4} \cmidrule(lr){5-7} 
                        & Acc. (↑)      & ECE (↓)      & NLL (↓)     & Acc. (↑)      & ECE (↓)      & NLL (↓)     \\ \midrule
Standard                & \textbf{0.9158}  & \textbf{0.0205}       & \textbf{0.2961}      & \textbf{0.7739} & \underline{0.0575}       & \textbf{0.9787}      \\
Single                  & 0.8965 & 0.0383       & 0.3791      & 0.7350 & \textbf{0.0525}       & 1.1667      \\
NDE (uni)               & \underline{0.9147} & \underline{0.0329}       & \underline{0.3408}      & 0.7151 & 0.0762       & 0.9975      \\
NDE (norm)              & 0.9132 & 0.0343       & 0.3453      & \underline{0.7660} & 0.0740       & \underline{0.9856}      \\ \bottomrule
\end{tabular}
\end{table}

\section{Ablation}
\label{sec:ablation}

\subsection{Effect of Different Weight Perturbation Strategies}
\label{subsec:ablation_weight_pertubation}
Here, we investigate the effects of varying the perturbation strategies applied to the weights. Specifically, we focus on two aspects: (i) the proportion of weights perturbed, denoted as \(\alpha\), and (ii) the scale of noise applied to the weights, denoted as \(\beta\). Details are discussed in Section \ref{sec:noisy_deep_ensemble}. Fig. \ref{fig:noise_scale} shows the test accuracy on CIFAR-10 when training the Noisy Deep Ensemble with different values of \(\alpha\) and \(\beta\). A uniform distribution is considered for the perturbation distribution.

As shown in Fig. \ref{fig:noise_scale}, the proportion of weights affected by perturbations has a minor impact on test accuracy. In contrast, the scale of the noise significantly influences test accuracy. Specifically, test accuracy declines when the noise scale is reduced to $\beta \leqq 0.1$. This decrease occurs because a smaller noise scale tends to converge towards the same local minima as the original \textit{parent model}, resulting in a loss of prediction diversity in ensemble learning. On the other hand, even when the noise scale is increased to $\beta \geqq 5.0$, the decrease in test accuracy is minimal. This is due to the retraining of each model after perturbing the weights. When $\alpha$ is set to 0, i.e., no weight perturbation, the test accuracy is lower than in most perturbed settings. This suggests that our proposed weight perturbation is essential.

\subsection{Scaling Number of Ensemble Members}
\label{subsec:scaling_ensemble_members}

\begin{figure}[t]
  \begin{minipage}{0.47\textwidth}
    \includegraphics[width=\linewidth]{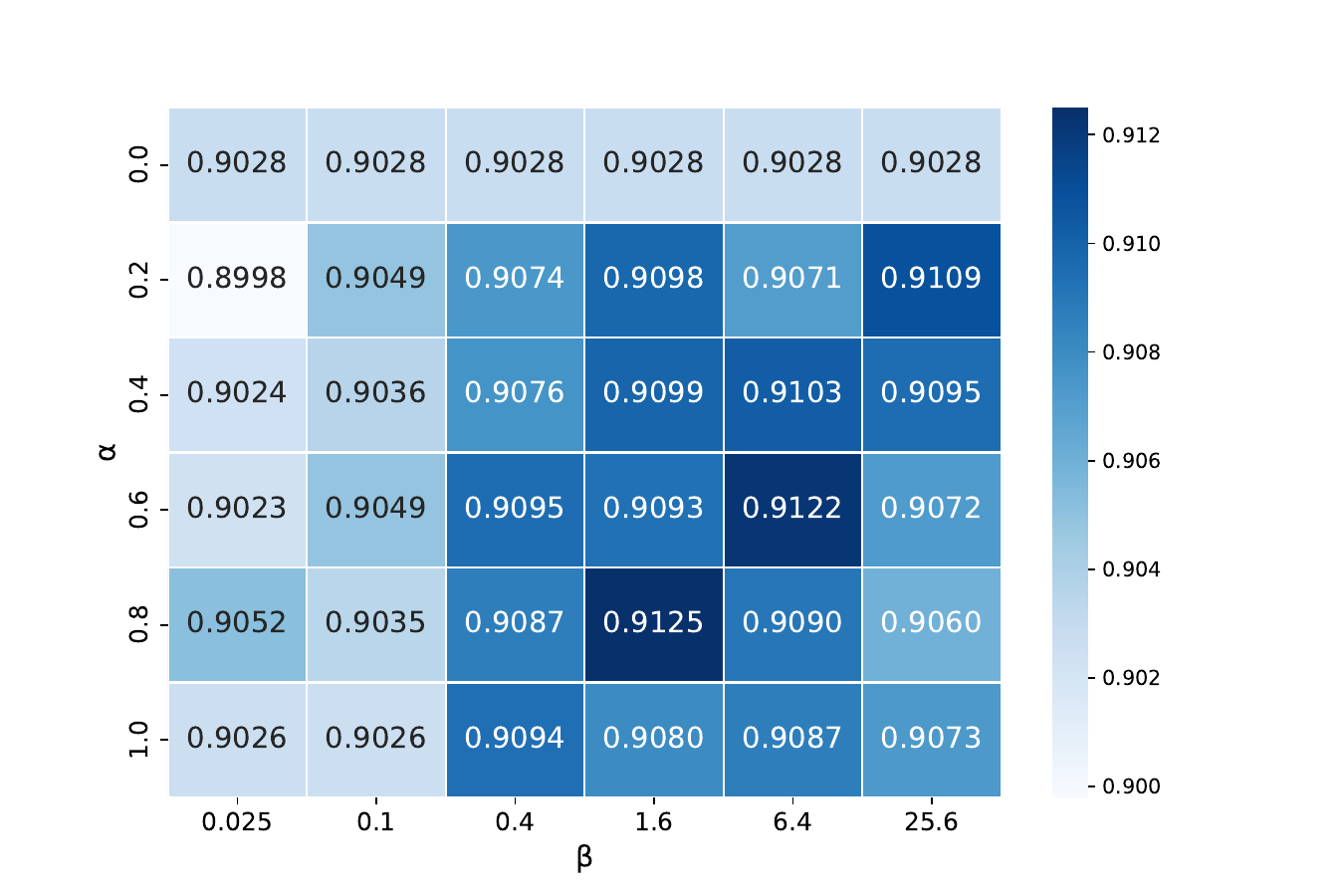}
    \caption{The effects of changes in perturbations on the accuracy}
    \label{fig:noise_scale}
  \end{minipage}%
  \hspace{1em}
  \begin{minipage}{0.47\textwidth}
    \includegraphics[width=\linewidth]{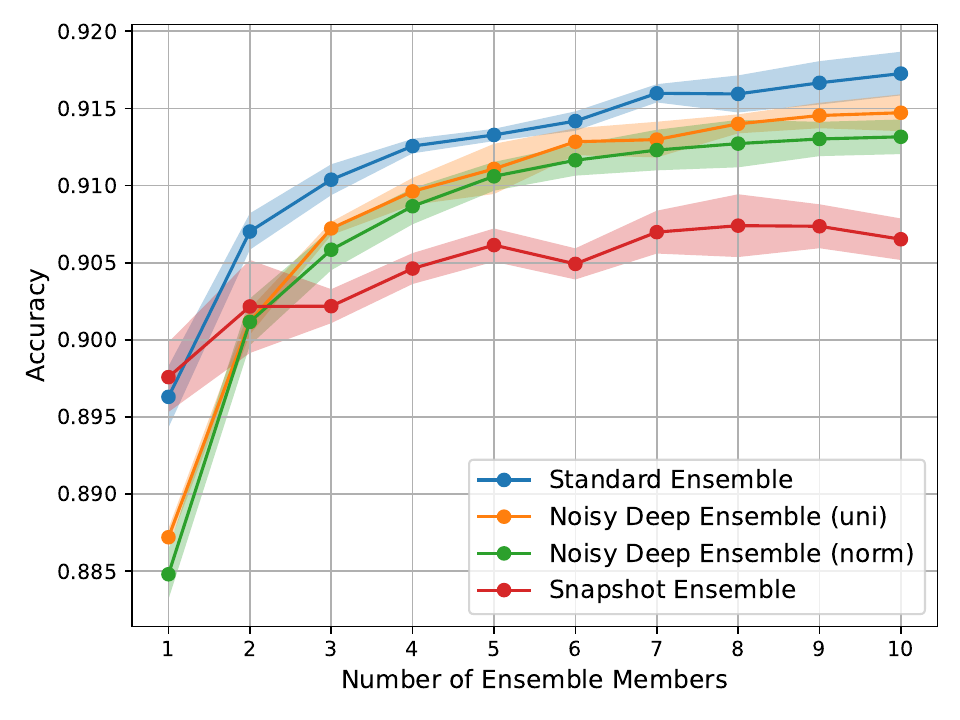}
    \caption{The effects of changes in the number of ensemble members on the accuracy}
    \label{fig:num_ensemble_members}
  \end{minipage}
\end{figure}

This section examines the impact of increasing the number of members in an ensemble. In standard ensemble learning, performance is typically improved as the number of ensemble members increases  \cite{WhyMH}. We investigate whether a similar trend is observed in the Noisy Deep Ensemble. The experimental results are shown in Fig. \ref{fig:num_ensemble_members}. Here, the Noisy Deep Ensemble was trained with varying numbers of ensemble members, and the mean and standard deviation of 5 trials are plotted. As shown in Fig. \ref{fig:num_ensemble_members}, the Noisy Deep Ensemble demonstrates improved test accuracy with increased ensemble members.
Conversely, the Snapshot Ensemble\cite{Snapshot} shows a plateau improvement in test accuracy despite increased ensemble members. This plateau is attributed to the lower diversity of predictions among ensemble members derived from a single model's learning process compared to those trained independently (Section \ref{subsec:evaluating_ensemble_effectiveness}). The Noisy Deep Ensemble exhibits scaling capabilities comparable to typical ensembles with uniform (uni) and Gaussian (norm) perturbation distributions.

\subsection{Performance with Various Optimisers}
We validate noisy deep ensemble is compatible with various optimizers such as RMSProp \cite{RMSProp}, AdamW \cite{AdamW}. The results are shown in Table \ref{tab:optimizer_performance}. Because we use the same hyperparameters tuned for SGD, the performances of RMSProp and AdamW are slightly worse. However, we can see similar performance improvement as SGD. 

\begin{table}[t]
\centering
\setlength{\tabcolsep}{6pt} 
\captionsetup{skip=2pt}
\caption{Comparison of optimization algorithms for different methods. The value in the table represents test accuracy on CIFAR-10.}
\label{tab:optimizer_performance}
\begin{tabular}{@{}lccc@{}}
\toprule
Method        & SGD & RMSProp & AdamW \\ \midrule
Standard      & \textbf{0.9158} & \textbf{0.8637} & \underline{0.8742} \\
Single        & 0.8965 & 0.8017 & 0.8194 \\
NDE (uni)     & \underline{0.9147} & \underline{0.8392} & \textbf{0.8744} \\
NDE (norm)    & 0.9132 & 0.8351 & 0.8554 \\ \bottomrule
\end{tabular}
\end{table}

\section{Conclusion}
In this study, we proposed the Noisy Deep Ensemble, a method designed to make the training of neural network ensembles more efficient. This method demonstrated superior test accuracy compared to conventional ensemble methods across various CNN architectures on CIFAR-10 and CIFAR-100. 

In some cases, Noisy Deep Ensemble performs worse than a single model (see Table \ref{tab:model_performance}). The uncertain nature of our perturbation process causes this. For example, the locations for weight perturbation were chosen randomly. From the perspective of neural network pruning, it has been shown that ignoring most weights selected appropriately has minimal impact on performance  \cite{OptimalBD}. Insights gained in this area could be utilized to refine the selection of weights to perturb in the Noisy Deep Ensemble, such as preferentially perturbing weights with larger absolute values. 

\begin{credits}
\subsubsection{\ackname} This work was supported in part by the Japan Society for the Promotion of Science (JSPS) KAKENHI Grant-in-Aid for Scientific Research (C) under Grant 23K11164. This work was also supported by several competitive funds within the University of Fukui.
\end{credits}


%
%
%
\bibliographystyle{splncs04}
\bibliography{mybib}

\end{document}